\documentclass[5p,times, ,twocolumn]{elsarticle}

\usepackage{lipsum}

\journal{Journal of \LaTeX\ Templates}

\usepackage[figuresright]{rotating}
\usepackage{graphicx}
\usepackage{slashbox}
\usepackage{xcolor}
\usepackage{multicol}
\usepackage{algorithm}
\usepackage{algorithmic}
\usepackage[none]{hyphenat}

\usepackage{array}
\usepackage{multirow}
\usepackage{amsmath,amssymb}

\usepackage{lipsum}

\usepackage{graphicx}
\usepackage{eepic}
\usepackage{epic}
\usepackage{rotate}
\usepackage{psfrag}
\usepackage{psboxit}

\def\vec#1{\mathchoice{\mbox{\boldmath  $\displaystyle\bf#1$}}
{\mbox{\boldmath  $\textstyle\bf#1$}}
{\mbox{\boldmath  $\scriptstyle\bf#1$}}
{\mbox{\boldmath  $\scriptscriptstyle\bf#1$}}}

\def\mat#1{\mathchoice{\mbox{\boldmath$\displaystyle\tt#1$}}
{\mbox{\boldmath$\textstyle\tt#1$}}
{\mbox{\boldmath$\scriptstyle\tt#1$}}
{\mbox{\boldmath$\scriptscriptstyle\tt#1$}}}

\begin{document}

\begin{frontmatter}

\title{SMC Faster R-CNN: Toward a Scene-Specialized Multi-Object Detector}

\author[rvt,abs]{Ala Mhalla}
\ead{mhallaala@gmail.com}
\author[abs]{Thierry Chateau}
\ead{thchateau@gmail.com}

\author[rvt,abs]{Houda Ma\^amatou}
\ead{maamatou.houda@gmail.com}

\author[rvt]{Sami Gazzah}
\ead{sami.gazzah@gmail.com}

\author[rvt]{Najoua Essoukri Ben Amara}
\ead{najoua.benamara@eniso.rnu.tn}

\address[rvt]{LATIS ENISo, National Engineering
School of Sousse, University of Sousse, Tunisia}
\address[abs]{Institut Pascal, Clermont Auvergne University, France}

\begin{abstract}
Generally, the performance of a generic detector decreases significantly when it is tested on a specific scene due to the large variation between the source training dataset and the samples from the target scene. To solve this problem, we propose a new formalism of transfer learning based on the theory of a Sequential Monte Carlo (SMC) filter to automatically specialize a scene-specific Faster R-CNN detector.
The suggested framework uses different strategies based on the SMC filter steps to approximate iteratively the target distribution as a set of samples in order to specialize the Faster R-CNN detector towards a target scene.
Moreover, we put forward a likelihood function that combines spatio-temporal information extracted from the target video sequence and the confidence-score given by the output layer of the Faster R-CNN, to favor the selection of target samples associated with the right label.
The effectiveness of the suggested framework
is demonstrated through experiments on several public traffic datasets. Compared with the state-of-the-art specialization frameworks, the proposed framework presents encouraging results for both single and multi-traffic object detections.
\end{abstract}


\end{frontmatter}


\section{Introduction}
Learning-based object detection algorithms have become an essential part for numerous video analysis applications, including security and intelligent transportation systems \cite{pan2010traffic}\cite{benfold2011stable}. However, most detectors are learnt with generic annotated datasets that are sampled from a large number of situations to cover the maximum intra-class variability of the traffic objects. When applied on a specific scene, the distribution of objects captured by the camera, like the Closed-Circuit Television camera (CCTV camera), is only a small subset of the initial learning set, and the resulting generic detector is often limited.
Therefore, the detector may fail to perform satisfactorily when tested on scenes that have data distributions different from the source training dataset \cite{dollar2009pedestrian}\cite{wang2009unsupervised}.

This problem can be solved by transfer learning, referred to as cross-domain adaptation, which can specialize a generic detector to a target scene. A classical way of specializing a generic detector is to manually select positive and negative samples from the target scene to re-train a scene-specific one. This requires collecting labelled data in every new scene and training a new detector, which can be labor intensive. A typical solution to avoid these tasks is to automatically label samples from the target scene and to transfer only a set of useful target samples to re-train a scene-specific detector.

Most state-of-the-art researches have been recently made to iteratively develop a scene-specific detector, whose training process is aided by generic detectors for automatically collecting training samples from target scenes without manually labelling them \cite{benfold2011stable}\cite{maamatou2016}\cite{wang2014scene}\cite{htike2014efficient}.
Accordingly, we put forward a new formalization of transfer learning based on the theory of a Sequential Monte Carlo (SMC) filter \cite{smith2013sequential} so as to automatically generate a specialized Faster R-CNN detector \cite{DBLP:journals/corr/RenHG015} for multi-traffic object detection, enhancing perform better than the generic one.

\begin{figure*}
\label{img1}
\centering
\includegraphics[width= 11cm,height=48mm]{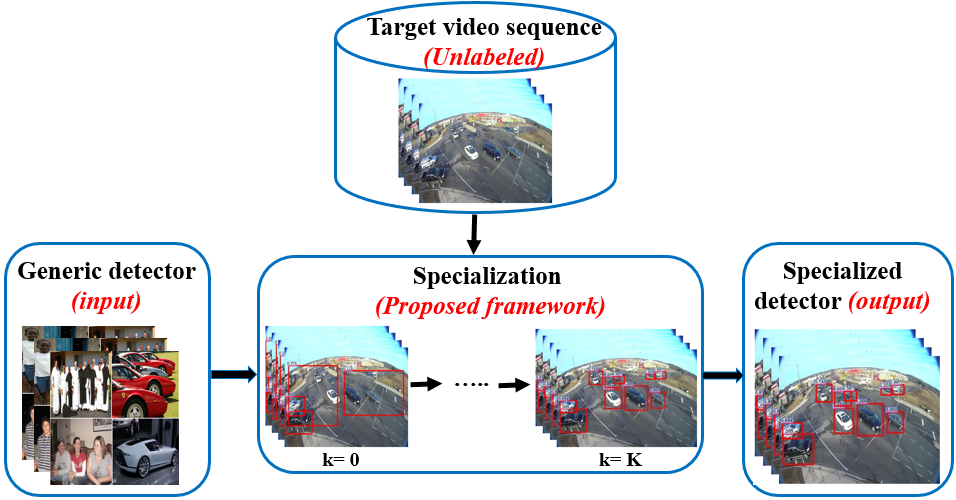} 
\\
(a)\\
\centering
\includegraphics[width= 13cm,height=44mm]{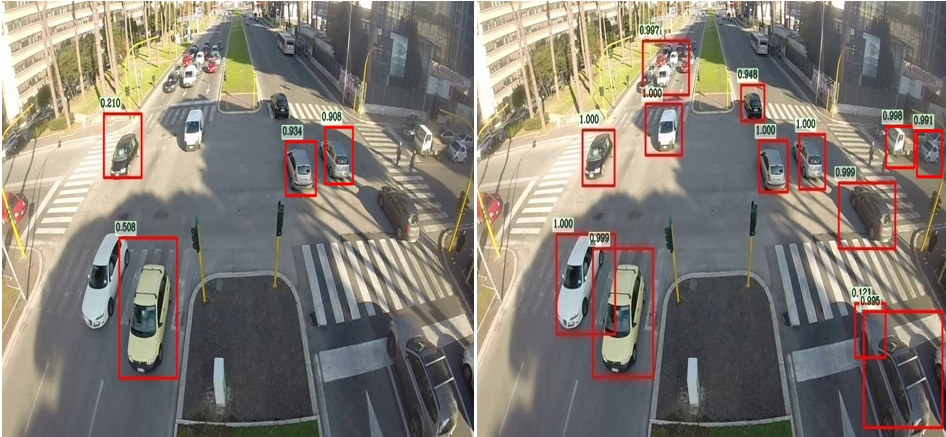}
\\
(b)\\
\centering
\includegraphics[width= 13cm,height=44mm]{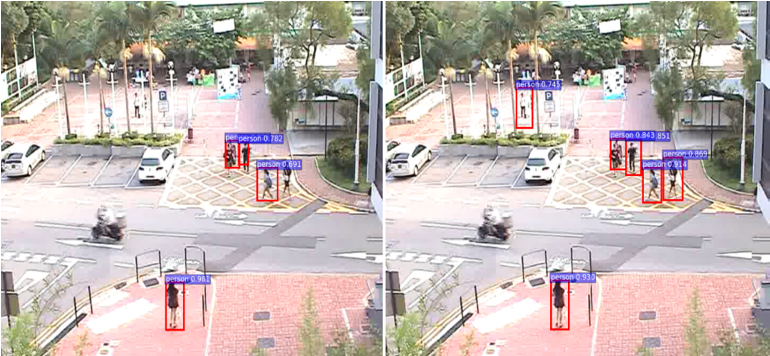}
\\
(c)\\
\centering
\includegraphics[width= 13cm,height=44mm]{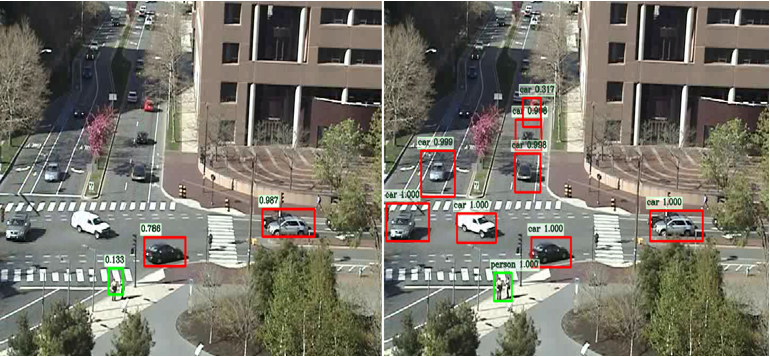}\\
(d)\\
\caption{\label{fig_synop1}(a) General synoptic of the proposed framework. The input of the framework is a generic Faster R-CNN detector fine-tuned on a generic dataset, then given a target video sequence without any label information, an iterative process automatically estimates both the set of target objects and the parameters to specialize the Faster R-CNN deep detector; (b) and (c) improvement of specialized scene-specific detector over generic detector for single-class and (d) multi-class object detection (left images are generic Faster R-CNN detections and right images are specialized Faster R-CNN ones).}
\end{figure*}

A global synoptic of our framework is illustrated in Figure \ref{fig_synop1}.(a). We have a generic Faster R-CNN detector which is fine-tuned by a source labelled dataset with labeled information given in the form of traffic-object
annotations. Given a target video sequence where labeled information is not available, an iterative process estimates both the set of target objects and the parameters of the specialized Faster R-CNN detector. This latter is automatically and iteratively trained and is called until a stopping criterion is reached. Then a final specialized Faster R-CNN detector is produced.

Our main contribution consists in putting forward a new transfer learning framework based on the formalism and the theory of the SMC filter for deep detector specialization.
The aim of our formalization is to automatically label the target data, to favor the selection of the target samples associated with the right label and to fine-tune a scene specialized Faster R-CNN detector. 

Although the use of the SMC filter for transfer learning is obviously not new, our work extends the SMC framework for deep detector and for multi-traffic object detection. Moreover, we propose new strategies for transfer learning inspired from the three steps of the SMC filter :

\textbf {(1) Strategy of bounding box proposals:} In order to use target samples for training a scene-specialized detector, the first strategy of the algorithm is to propose bounding boxes of traffic-object candidates by adapting the architecture of the Faster R-CNN deep network for only traffic-object detection.
This strategy gives a set of suggestions composed by traffic proposals predicted by the output layers of the Faster R-CNN.

\textbf {(2) Strategy of verification:} We suggest a verification strategy to correctly select unlabeled samples from a target scene. This strategy utilizes a combination between the confidence-scores
returned by the output layer of the Faster R-CNN and the visual context cues extracted from the target video sequence, in order
to favor the selection of positive samples from a target scene and to reduce the risk of introducing wrong labelled examples in the training dataset.


\textbf {(3) Strategy of sampling:} 
We suggest a sampling strategy that collects useful samples from target datasets according to their weights importance, reflecting the likelihood that they belong to the target distribution. The main role of this strategy is to build the specialized dataset with samples produced by the strategy of verification. 
To do this, we use the Importance Sampling (IR) algorithm inspired from the theory of the SMC filter \cite{doucet2001}. This algorithm transforms the weight on a number of repetitions, through repeating the samples associated to a high weight by numerous ones and repeating the samples associated to a low weight by few ones.
This strategy makes the suggested framework
applicable to specialize any detector and avoids the distortion of the specialized dataset, while selecting training samples according to the importance of their weights without modifying the training function. 



Another contribution is to make a comparative evaluation of the proposed framework to the state-of-the-art specialization frameworks on several public datasets and with new more challenging annotations.

The rest of the paper is organized as follows. Section 2 reviews the existing work performed in this field and provides a discussion about the advantages of our work over the state-of-the-art specialization frameworks. After that, a detailed description of our approach are provided in section 3. The experiments and results are described in section 4. Finally, the conclusion is given in section 5.

\section{Related Work}

\subsection{\textbf {State-of-the-art scene specialization frameworks}}

In this subsection, we are interested in the related specialization frameworks that suggest to automatically specialize scene-specific detectors or classifiers towards a target scene.

In the recent years, transfer learning has attracted a lot of research groups in developing state-of-the-art theories and new applications in several domains like object detection and recognition \cite{maamatou2016}\cite{wang2014scene}\cite{htike2014efficient}\cite{mao2015}.
Transfer learning aims to address the problem when the distribution of the training data from the source domain is different from that of the target one. 

According to the state-of-the-art theories, transfer learning approaches suggest to use the available annotated data and knowledge acquired through some previous tasks relative to source domains so as to improve a learning system of a target task in a target domain.

Generally, three categories of transfer learning methods, related to the proposed framework, were described in \cite{maamatou2016}.
The first one would change the parameters of a source learning model to improve its accuracy in a target domain \cite{tommasi2014learning}\cite{aytar2011tabula}. The second category would decrease the variation between the source and target distributions to adapt a detector to the target domain \cite{pan2011domain}\cite{quanz2012knowledge}. The third one would automatically choose the training samples that could provide a better detector or classifier for the target task \cite{maamatou2016}\cite{mao2015}.
In this paper, we focus on the third category which utilizes an automatic labeler to select data from the target domain.

Much of the state-of-the-art research has used an iterative self-training process to specialize a generic detector to a target scene. An ideal framework can apply a generic detector on some frames in a target scene, score each detection using some heuristics and then include the most confident positive and negative detections to the original dataset for retraining \cite{wang2012transferring}\cite{rosenberg2005semi}\cite{levin2003unsupervised}. 
Rosenberg {\it {\it {\it et al.}}} \cite{rosenberg2005semi} opted for a self-training framework based on background subtraction to label scene samples. Only the samples with high confidence scores were added in a new training dataset from one iteration to another. Contrarily, there was a risk of introducing a wrong labelled example in the training dataset, which may degrade the framework performance over iterations. In addition, Wang {\it {\it {\it et al.}}} \cite{wang2014scene} utilized different contextual cues such as visual appearances of objects, motion of pedestrian, model of road, size and location to select positive and negative samples from the target scene and to add the last ones in the training dataset for retraining. This approach proved to be sensitive to the risk of drifting and it can be applied only onto a particular classifier.

Moreover, some solutions collected the training source dataset with new samples extracted from the target scene, which increased the time of training and the size of the dataset during iterations \cite{aytar2011tabula}\cite{quanz2012knowledge}. 
Others were limited only to the use of samples extracted from the target domain \cite{all2011flowboost}\cite{mao2015}, which caused the loss of useful samples stored in the source dataset.
Htike {\it {\it {\it et al.}}} \cite{htike2014efficient} presented an approach that used only target samples labeled by a background subtraction algorithm and verified by the tracklet method to train a specific detector. In the same vein, Mao and Yin \cite{mao2015} used tracklet chains to automatically label target information. They associated the proposal samples predicted by an appearance-object detector
into tracklets and they propagated labels to uncertain tracklets based on a comparison between their features and those of labeled tracklets. This framework used many manual parameters and several thresholding rules for every target scene, which can affect the specialization performance. 

Other solutions were proposed in \cite{maamatou2016}\cite{li2015domain}\cite{wang2012transferring}, which collected new samples from the target scene and the source dataset.
Maamatou {\it {\it {\it et al.}}} \cite{maamatou2016} suggested a transfer learning method based on the SMC filter to iteratively build a new specialized dataset that was used to train a new specialized pedestrian detector. This produced dataset consisted of both source and target samples that were utilized to estimate the unknown target distribution. Our proposed framework is inspired from this latter.

Addressing this problem with deep learning has recently gained a growing attention. Some deep models have been investigated in the unsupervised and transfer learning challenge \cite{guyon2011unsupervised}. 
Transfer learning using deep models has been turned out to be effective in some challenges \cite{mesnil2012unsupervised}\cite{goodfellow2012spike} like traffic-object detection \cite{zeng2014deep}\cite{li2015domain}, emotion recognition \cite{ng2015deep} and sentiment analysis \cite{glorot2011domain}.
In order to take advantage of these types of detectors, several transfer learning methods have been proposed to specialize a Convolutional Neuronal Network (CNN) detector by fine-tuning an ImageNet-pre-trained model with a small target dataset.
Li {\it {\it {\it et al.}}} \cite{li2015domain} suggested adapting a generic CNN vehicle detector to a target scene by appropriating the shared filters between source and target data and updating the non-shared filters. In contrary to \cite{li2015domain}\cite{oquab2014learning}, which needed several manual labeling of data in the target scene, Zeng  {\it {\it {\it et al.}}} \cite{zeng2014deep} proposed to use Wang's approach \cite{wang2014scene} to select target samples and utilized these latter as an input to their CNN deep model to re-weight samples from target and source domains without manually  labeling data from the target scene.

In this paper, we use a recent deep model, the Faster R-CNN \cite{DBLP:journals/corr/RenHG015}, thanks to its efficiency and robust performance in general object detection and we specialize it with a new formalism of transfer learning based on the theory of the SMC filter \cite{smith2013sequential} for multi-traffic object detection.

The Faster R-CNN was put forward in \cite{DBLP:journals/corr/RenHG015} to accurately detect general objects in pictures. It achieved a state-of-the-art 73.2 mean average precision on the PASCAL VOC 2007 dataset. 
It was composed of two modules: 
The first module is a Region Proposal Network (RPN) that provided a set of rectangular object proposals from an input image. The second module was the Fast R-CNN deep model \cite{girshick2015fast} which took as inputs this set of object proposals and then used them for classification.
The entire system was a single, unified network for object detection.  

The suggested framework presented in this paper proposes some improvements over the related specialization frameworks. These improvements will be described in the next subsection.

\subsection{\textbf {Literature analysis and framework proposition}}

This section provides a discussion about the advantages of our work over the state-of-the-art scene specialization frameworks and the main difference between the SMC framework proposed by Maamatou {\it {\it {\it et al.}}} \cite{maamatou2016} and the suggested one.

Most of the specialization frameworks cited above are based on hard-thresholding rules and are very sensitive to the risk of drifting during iterations, or they are applied only to particular classifiers or few detectors like the HOG-SVM. In fact, several frameworks are limited only for mono-traffic object detection, or they need many iterations for the convergence of the specialization process.

Differently from the existing work, we put forward an iterative process based on the formalism of the SMC filter to specialize the Faster R-CNN deep detector for multi-traffic object detection. Accordingly, our proposed framework allows reducing the risk of drifting by using efficient strategies during iterations and it can be used to specialize any deep detector like the Fast R-CNN \cite{girshick2015fast} and the R-CNN \cite{Girshick_2014_CVPR}. Furthermore, this framework may be applied using several strategies on each step of the SMC filter. 
Particularly, we cite some advantages of the suggested framework: 
\begin{itemize}

\item  We propose a likelihood function based on an efficient strategy of verification. This latter is used to favor the selection of samples associated to the right label from a target scene,  
to decrease the risk of drifting the detector over iterations by reducing the introduction of mislabeled examples in the training dataset.

\item The suggested framework automatically specializes a generic detector to a target scene. This framework iteratively estimates the unknown target distribution as a specialized dataset by selecting only relevant samples from the target dataset. These samples are selected to re-train a specialized detector that increases the detection accuracy in the target scene. Contrarily, several state-of-the-art frameworks have aimed to collect samples from both source and target datasets to improve accuracy by augmenting the training dataset. These frameworks have led to extend the size of the training dataset and to slightly decrease the performance of the detector during iterations.

\item To permit training an accurate specialized detector with the same function as the generic one and avoiding the distortion of the specialized dataset, we suggest a sampling strategy 
which uses the IR algorithm to select the confidence samples relevant to their weight returned by the likelihood function. 
This makes our framework applicable to specialize any deep detector, while training the treating samples according to the importance of their weight without modifying the training function, as done by \cite{wang2014scene} \cite{wang2012transferring}.

\item We derive a generic transfer learning framework in which many strategies can be integrated in the SMC steps.

\end{itemize}

Table \ref{timep10} provides a comparison over the SMC framework proposed by Maamatou  {\it {\it {\it et al.}}}\cite{maamatou2016} and our suggested one.

\begin{table*}[!ht]

\caption{\label{timep10} Description of the difference between the work of Maamatou {\it {\it {\it et al.}}} \cite{maamatou2016} and our proposed one}
\centering
\begin{tabular}{l|l|l}
\centering
\backslashbox & Maamatou  {\it {\it {\it et al.}}} \cite{maamatou2016}  & Our framework \\
\hline
Generic detector & HOG-SVM & Faster R-CNN\\ 
Transfer learning & Positive \& negative samples& Positive samples\\
Specialized dataset & Source \& target samples & Target samples \\
Output & Specialized SVM & SMC Faster R-CNN \\
Specialized process & SMC steps & SMC steps \& fine-tuning step \\
\hline
Traffic objects & Pedestrian & Multi-traffic object \\
\end{tabular}
\end{table*}

The advantages of our specialization framework over the SMC framework \cite{maamatou2016} are:

\begin{itemize}

\item In \cite{maamatou2016}, for each iteration, they selected relevant samples from both source and target domains to create a specialized dataset. In contrast, our proposed framework selects only the relevant samples from target domains according to the importance of their weights to create a specialized dataset. This solution enables a faster learning
of detector and leads to an increase in detection accuracy.
\item The specialized framework proposed in \cite{maamatou2016} was very sensitive to the risk of drifting because they used only a background subtraction algorithm to assign weights to the target samples. Indeed, several static objects or those with similar background appearances were classified as negative samples, and mobile background objects were labeled as objects of interest. On the other hand, to avoid the distortion of the specialized dataset with mislabeled samples, we propose a likelihood function based on the verification strategy, which combines the confident-score given by the output layer of the Faster R-CNN network with spatial-temporal cues in order to attribute confidence weights to target samples.

\item The work of Maamatou {\it {\it {\it et al.}}}\cite{maamatou2016} was limited for only single-traffic object detection, but our proposed one is extended for multi-traffic objects like cars, pedestrians, buses, motorbikes...

\item Differently from the work in \cite{maamatou2016}, we put forward new strategies for transfer learning inspired from the three steps of the SMC filter to specialize the Faster R-CNN deep detector.

\item It is important to say that we need only two iterations for the convergence of our specialization process, whereas the framework suggested in \cite{maamatou2016} required at least 4 iterations for this convergence.

\item The proposed approach in \cite{maamatou2016} was limited to specialize the SVM classifier, in contrary, our framework is applicable to specialize some deep detector like the Fast R-CNN \cite{girshick2015fast}, the Faster R-CNN \cite{DBLP:journals/corr/RenHG015} and the R-CNN \cite{Girshick_2014_CVPR}.

\end{itemize}
\section{\textcolor{black}{Proposed specialization framework}}

\begin{figure*}
\centering
\label{img2}
\includegraphics[width=13.5cm]{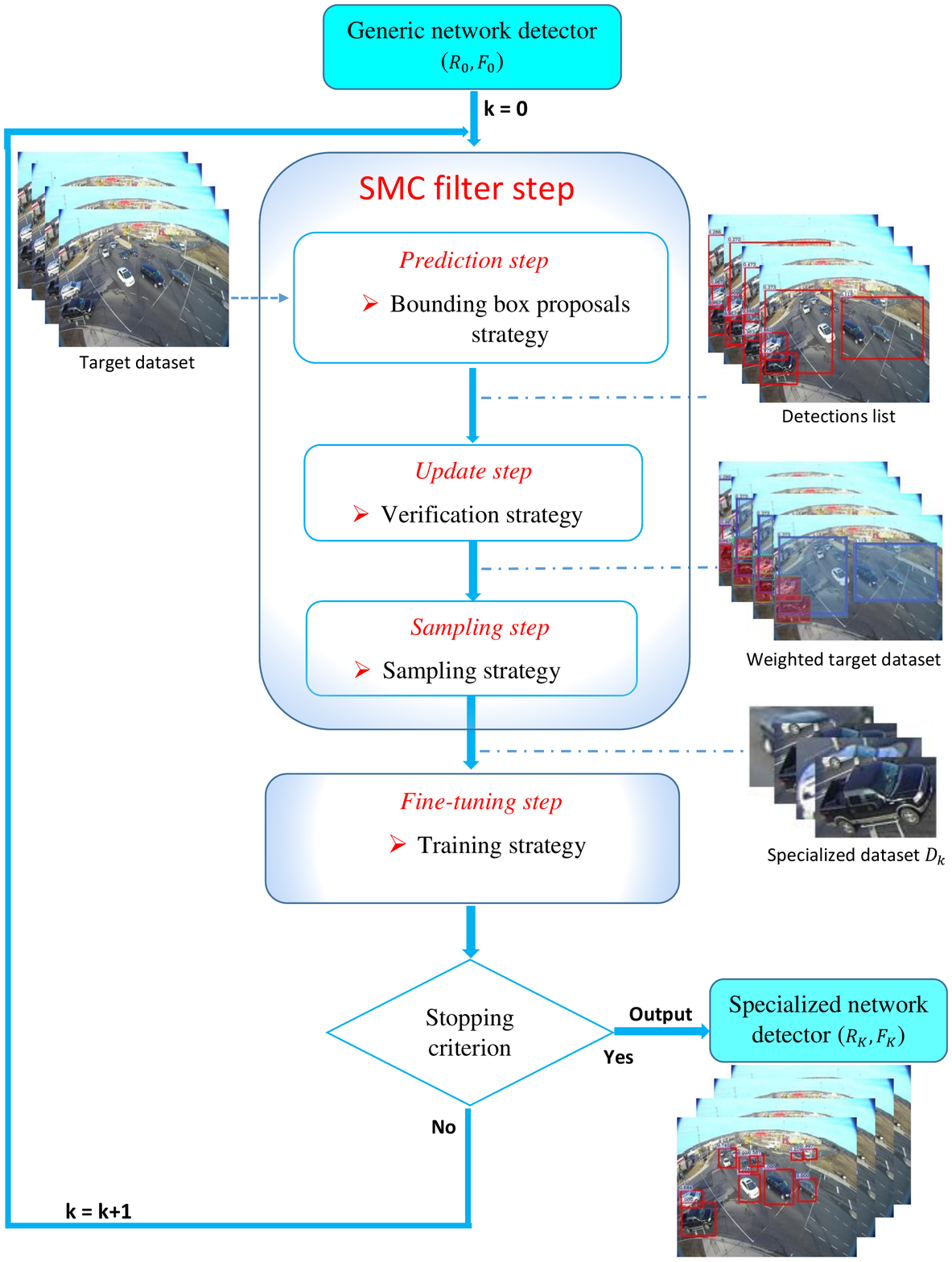}

\caption{\label{fig:global}Block diagram of proposed approach: At the first iteration, our generic detector $({\cal {R}}_{0},{\cal {F}}_{0})$ which is fine-tuned by the source dataset is utilized in the first prediction step by using bounding box proposals strategy to produce a list of traffic-object bounding boxes from the target scene, and then a update step based on the likelihood function is used to favor the selection of positive samples from a target scene. The sampling step determines which samples will be included in the specialized dataset by using the sampling strategy. A new specialized detector $({\cal {R}}_{k},{\cal {F}}_{k})$ is fine-tuned by utilized training strategy in the fine-tuning step, which will become the input of the prediction step in the next iteration $k= k+1$. A final specialized detector $({\cal {R}}_{K},{\cal {F}}_{K})$ is called when a predefined number of iterations is reached. The red rectangles in the output image of update step mean that samples have a high weights attributed by our suggested likelihood function and a blue ones mean that samples have a low weights.}
\end{figure*}

In this section, we present the proposed framework for specializing the Faster R-CNN model to a target scene based on SMC filter steps.
Figure \ref{fig:global} shows the block diagram representation corresponding to one iteration of our suggested SMC Faster R-CNN. First, a generic Faster R-CNN network $({\cal {R}}_{0},{\cal {F}}_{0})$ is fine-tuned on a generic dataset (eg: PASCAL VOC).
Given the videos taken by a stationary camera in target scenes, at a first iteration ($k$ = 1),
the generic detector $({\cal {R}}_{0},{\cal {F}}_{0})$ is applied in the prediction step by using the strategy of bounding box proposals to suggest a set of traffic-object proposals in each individual image.
Then an update step based on the likelihood function is used to favor the selection of the positive samples from a target scene by associating weight to each proposal sample returned by the prediction step. By utilizing the sampling strategy, the sampling step determines
which samples should be included in the specialized dataset according to their weights. A new specialized detector $({\cal {R}}_{k},{\cal {F}}_{k})$ is trained by using the training strategy in the fine-tuning step. This specialized one will become the input of the prediction step in the next iteration. The scene-specific detector is automatically and iteratively trained and is called until reaching a stopping criterion, for example a fixed number of iterations. When the number of iterations is reached, a final specialized detector $({\cal {R}}_{K},{\cal {F}}_{K})$ will be generated.

In what follows, we first describe the specialization of the Faster R-CNN model based on the theory of the SMC filter.

\subsection{\textbf {Faster R-CNN specialization based on SMC filter}}

Given a source dataset, from which a generic Faster R-CNN detector can be trained from this source dataset, and a video sequence of a target scene, then a specialized Faster R-CNN detector will be generated. This latter is the output of the distribution approximation provided by the formalism of the SMC filter and the fine-tuning step. To do this, let us define:
\begin{itemize}
    \item ${\cal I}_t\doteq \{\mat{I}^{(i)}\}_{i=1}^{I_i}$ is a set of unlabelled images extracted uniformly from a video sequence of a target scene. 
	\item ${\cal D}_k \doteq \{ \vec{x}_k^{(n)}\}_{n=1}^{N_k}$ is a specialized dataset at iteration $k$, where $\vec{x}_k^{(n)}$ is a target object sample to be detected in each target image of the set ${\{\mat{I}^{(i)}\}_{i=1}^{I}}$. This sample is defined by: $\vec{x}_k^{(n)} \doteq\{\vec{p}^{(n)}_k,y^{(n)}_k,s^{(n)}_k\}$ where $\vec{p}^{(n)}_k \doteq \{ u^{(n)}_k,v^{(n)}_k,w^{(n)}_k,h^{(n)}_k\}$ is the position of an object, with $(u^{(n)}_k,v^{(n)}_k)$ being the upper left coordinates of the object bounding box and $(w^{(n)}_k,h^{(n)}_k)$ being the width and the height of the object bounding box, $y^{(n)}_k$ is the object class label and $s^{(n)}_k$ is an associated score.
	\item $\{ \vec{x}^{(n)}\}_{n=1}^{N} = \Theta(\{\mat{I}^{(i)}\}_{i=1}^{I_i};{\cal {R}},{\cal {F}})$ is a function that applies the Faster R-CNN detector using the RPN network model ${\cal R}$ for the localization task and the Fast R-CNN network model ${\cal F}$ for detection. For both localization and detection, a set of candidate objects with associated scores is provided.

	\item $\{{\cal \tilde{R}},{\cal \tilde{F}}\} = f(\{\mat{I}^{(i)}\}_{i=1}^{I_i},\{ \vec{x}^{(n)}\}_{n=1}^{N};{\cal {R}},{\cal {F}})$ is a fine-tuning function that returns the new parameters ${\cal \tilde{R}}$ and ${\cal \tilde{F}}$ of the Faster R-CNN network. The fine-tuning is performed from the Faster R-CNN network with initial ${\cal {R}}$ parameters for the RPN and initial ${\cal {F}}$ parameters for the Fast R-CNN, utilizing a training dataset given by the set of images $\{\mat{I}^{(i)}\}_{i=1}^{I_i}$ and the associated objects $\{ \vec{x}^{(n)}\}_{n=1}^{N}$
	
\end{itemize} 
We define $\vec{x}_k$ to be a hidden random state vector associated to a joint distribution between labels and features of dataset samples at an iteration $k$ and $\vec{z}_k$ a random measure vector associated to information extracted from the target scene (i.e. visual spatio-temporal information). Based on our assumption, the target distribution can be approximated by iteratively applying equation (\ref{eq:bayes}):
\begin{equation}
	\label{eq:bayes}
	p(\vec{x}_{k}|\vec{z}_{0:k})=
	C. p(\vec{z}_{k}|\vec{x}_{k})\int_{\vec{x}_{k-1}}p(\vec{x}_{k}|\vec{x}_{k-1})p(\vec{x}_{k-1}|\vec{z}_{0:k-1})d\vec{x}_{k-1}
\end{equation}
where $C$ is a normalisation factor: $C=1/p(\vec{Z}_{k}|\vec{Z}_{0:k}$).

The SMC filter estimates the probability distribution $p(\vec{x}_k|\vec{z}_k)$ by a set of $N$ particles (samples in this case), according to equation  (\ref{eq:BT}):
\begin{equation}
  \label{eq:BT}
	p(\vec{x}_k|\vec{z}_k) \approx \sum_{n=1}^N {\pi}^{(n)}_{k} 
	\delta_{\vec{x}_k^{(n)}}(\vec{x}_k) 
\end{equation}

\begin{itemize}
  \item $\delta$  represents the Dirac function (\ref{eq:fct15}):
  
\begin{equation}
\delta_{\vec{x}_k^{(n)}}(\vec{x}_k)=
\left\lbrace
\begin{array}{ccc}
1  & \mbox{if} &  \vec{x}_k = \vec{x}_k^{(n)} \\
0 &   & otherwise\\
\end{array}\right.
\label{eq:fct15}
\end{equation}

\item ${\pi}^{(n)}_{k} \in [0,1] $ is the weight associated to sample $n$ at iteration $k$ and $N$ is the number of target samples (\ref{eq:poids1}):
\begin{equation}
\pi^{n}_k=\dfrac{\pi^{n}_{k-1}~p(\vec{z}_k|\vec{x}_k=\vec{x}_k^n)}{\sum_{n=1}^{N}\pi^{n}_{k-1}~p(\vec{z}_k|\vec{x}_k=\vec{x}_k^n)}
\label{eq:poids1}
\end{equation}
It is important to note that the sum of the weights of all the samples is equal to (\ref{eq:fct12}):
\begin{equation}
\sum_{n=1}^N {\pi}^{(n)}_{k} = 1
\label{eq:fct12}
\end{equation}

\end{itemize}

All notations mentioned above are introduced in \cite{smith2013sequential}.

Therefore, the formalism of the SMC filter is used to approximate the unknown joint distribution of traffic objects by a set of samples that are initially unknown. We suppose that the iterative process selects relevant samples for the specialized dataset from one iteration to another, leading to converge to the right target distribution, and making the
resulting Faster R-CNN detector more and more efficient.

The resolution of equation (\ref{eq:bayes}) is divided into three steps: prediction, update and sampling. These steps are similar to the popular particle filter framework, widely used to solve the tracking problems in computer vision \cite{mei2011robust}\cite{smal2007advanced}. The details of
the three main steps are described in the following
subsections.
\subsubsection{Prediction step}
The prediction step consists in applying the Chapman-Kolmogorov equation (\ref{eq:chapman}):
\begin{equation}
	p(\vec{x}_{k}|\vec{z}_{0:k-1}) = \int_{\vec{x}_{k-1}} p(\vec{x}_{k}|\vec{x}_{k-1}) p(\vec{x}_{k-1}|\vec{z}_{0:k-1})d\vec{x}_{k-1}
	\label{eq:chapman}
	\end{equation}

Equation (\ref{eq:chapman}) uses the system dynamics term $p(\vec{x}_{k}|\vec{x}_{k-1})$ between two iterations in order to suggest a specialized dataset ${\cal D}_k \doteq \{ \vec{x}_k^{(n)}\}_{n=1}^{N_k}$ producing the
approximation (\ref{estimation}):

\begin{equation}
\label{estimation}
	p(\vec{x}_{k}|\vec{z}_{0:k-1}) \approx \{\tilde{\vec{x}}_{k}^{(n)}\}_{n=1}^{\tilde{N}_{k}}
\end{equation}
We suggest to extract the proposal set $\{\tilde{\vec{x}}_{k}^{(n)}\}_{n=1}^{\tilde{N}_{k}}$ from the set of proposals produced by the Faster R-CNN fine-tuned  by  
$\{ \vec{x}_{k-1}^{(n)}\}_{n=1}^{N_{k-1}}$ (the target set at iteration $k-1$):
\begin{equation}
	\{\tilde{\vec{x}}_{k}^{(n)}\}_{n=1}^{\tilde{N}_{k}} = \Theta(\{\mat{I}^{(i)}\}_{i=1}^{I_i};{\cal {R}}_{k-1},{\cal {F}}_{k-1})
\end{equation}
with a first iteration ($k=1$) that uses an initial generic network $({\cal {R}}_{0},{\cal {F}}_{0})$.

\subsubsection{Update step}

This step defines the likelihood term (\ref{eq:likelihood}) by utilizing a likelihood function. This latter assigns a weight $\tilde{\pi}$ to each proposal sample $\{\tilde{\vec{x}}^{(n)}_{k}\}_{n=1}^{\tilde{N}_{k}}$ returned by the Faster R-CNN at the prediction step.

\begin{equation}
   \label{eq:likelihood}
	p(\vec{z}_{k}|\vec{x}_{k}=\tilde{\vec{x}}^n_{k}) \propto \tilde{\pi}^n_{k}
\end{equation}

The likelihood function employs visual contextual cues extracted from the target video sequence and the confidence scores given by the output layer of the Faster R-CNN, to attribute a weight for each sample. More details about the likelihood function are given in section \ref{like}. The update step gives as an output a set of weighted target samples, which will be referred to as "the weighted target dataset" hereafter (\ref{basep}):
\begin{equation}
\label{basep}
\lbrace({\tilde{\vec{x}}}^{(n)}_{k}, \tilde{\pi}^{(n)}_{k}) \rbrace_{n=1}^{\tilde{N}_k}
\end{equation}
where $\{\tilde{\vec{x}}^{(n)}_{k},\tilde{\pi}^{(n)}_{k} \}$ represents a target sample with its associated weight and ${\tilde{N}_k}$ is the number of weighted samples.


\subsubsection{Sampling step}
The aim of this last recursive-filter step is to build a new specialized dataset by deciding, according to the strategy of sampling (defined in the contribution), which samples will be included in the produced dataset
${\cal D}_{k}=\{\vec{x}_{k}^{(n)}\}_{n=1}^{N_{k}}$ at the iteration $k$ from the weighted dataset $\{\tilde{\vec{x}}^{(n)}_{k},\tilde{\pi}^{(n)}_{k} \}_{n=1}^{\tilde{N}_k}$.
A sampling strategy is applied in order to generate a new unweighted dataset which has the same number of samples as the weighted one. To do this, we apply the IR algorithm, according to equation (\ref{eq:BTS}):
\begin{equation}
    \label{eq:BTS}
	{\cal D}_k=\{{\vec{x}}^{(n)}_{k} \}_{n=1}^{N_{k}} = IR\left(\{\tilde{\vec{x}}^{(n)}_{k},\tilde{\pi}^{(n)}_{k}\}_{n=1}^{\tilde{N}_{k}}\right)
\end{equation}
This step generates a new set ${\cal D}_{k}$ by drawing samples according to the weight $\tilde{\pi}^{(n)}_{k}$

\subsection{\textbf{Likelihood function}}
\label{like}

In order to choose the correct proposal, we put forward a likelihood function based on the verification strategy, which assigns a weight $\pi^{(n)}_k$ for each sample $\tilde{\vec{x}}^{(n)}_k$ returned by the prediction step. Our specifically designed likelihood function not only incorporates the confidence scores given by the output layer of the Faster R-CNN but also adds a spatial-temporal cues, to prioritize the selection of the correct samples and to reduce the risk of including wrong proposal samples in the specialized dataset.

Summarising the tests carried out on different databases, it is noticed that the generic Faster R-CNN is robust to generate true positive samples with a high score, and its selection of these ones will start to fail when the score of samples is lower than the score threshold $\alpha_k$. For this reason, we keep the samples which have a confidence score greater than or equal to $\alpha_k$ and we propose an observation function $f_L$ to assign a weight to each proposal sample that has a score lower than $\alpha_k$, according to (\ref{fct}):

\begin{equation}
\pi^{(n)}_k=
\left\lbrace
\begin{array}{ccc}
s^{(n)}_k  & \mbox{if} &s^{(n)}_k \geq \alpha_k\\
f_L(\tilde{\vec{x}}^{(n)}_k) & \mbox{if}  &s^{(n)}_k < \alpha_k\\
\end{array}\right.
\label{fct}
\end{equation}

Accordingly, we choose a dynamic threshold through iterations to avoid the problem of integrating negative samples into the specialized dataset. We are not limited to a fixed predefined threshold because the choice will be dynamic and will be related to the following equation (\ref{fct1}):
\begin{equation}
\alpha_k=
\left\lbrace
\begin{array}{ccc}
\cfrac{\tilde{s}_k}{\tilde{s}_{k-1}} \ \alpha_{k-1} & \mbox{if}  & {k}\ne 0 \\
\alpha_0 & \mbox{if}  & {k}= 0 \\
\end{array}\right.
\label{fct1}
\end{equation} 

where $\alpha_0$ is the initial value of the score threshold (fixed to $0.5$ for our experiments) and $\tilde{s}_k $ is the mean value of $s_{k}^{(n)}$ at iteration $ k $ :

Lower than $\alpha_k$, the deep detector will start to fail and it will become unable to correctly select positive samples from a specific scene. To solve this problem, we propose an observation function $f_L$ in order to favor the selection of positive samples using the information extracted from the target scene.
This function is based on the visual spatio-temporal cue "Background extraction overlap score", to attribute a weight for each sample.

In a traffic scene, it is rare for a traffic object to stay fixed for a long time, and a good detection occurs on a foreground blob; whereas, false positive background detections give some Region of Interests (RoIs) that appear over time at the same location and with the same size. 

\begin{figure}
\label{img41}
\centering
\includegraphics[width=0.8\columnwidth]{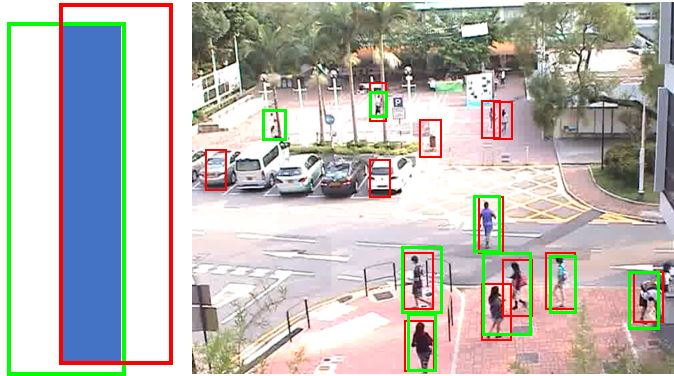}
\caption{\label{fig:fine11} 
The red rectangle presents the area in pixels of the considered RoI, the green rectangle is the foreground area, and the rectangle filled in blue is the area of intersection.}
\end{figure}

To assign a weight for each sample, we calculate an overlap\_score $\lambda_o$ (equation \ref{overlap}) that compares the RoI associated to one sample with the output of a binary foreground extraction algorithm.

\begin{equation}
\lambda_o \doteq \dfrac{2(RoI\_AR\ \times\ FG\_AR)}{RoI\_AR + FG\_AR}
\label{overlap}
\end{equation}
where $RoI\_AR$ is the area in pixels of the considered RoI and $FG\_AR$ is the foreground area at the RoI position (see Figure \ref{fig:fine11}).

The background subtraction algorithm used in the proposed observation function is adopted from \cite{zivkovic2006efficient} and was called the $"BackgroundSubtractorMOG2"$ algorithm. This latter is a Gaussian mixture-based background / Foreground segmentation algorithm.

One important property of this algorithm is that it chooses the appropriate number of Gaussian distribution for each pixel. It provides better adaptability to illumination changes. 
In our work, to ameliorate the result generated by the background subtraction algorithm mentioned above, we put forward some improvements such that:
\begin{itemize}
\item We apply several morphological filtering operations like erosion and dilation to the result of this algorithm so as to remove unwanted noise.

\item We remove the blobs which have a surface area less than 100 pixels.

\end{itemize}


The observation function (Algorithm \ref{alg_pos}) will assign a high weight to a positive proposition if it has an overlap\_score $\lambda_o$ that exceeds a fixed threshold $\alpha_p$, which is determined empirically.
\begin{algorithm}
\caption{Observation function}
\label{alg_pos}
\begin{algorithmic}
\REQUIRE 
{ Set $\{\tilde{\vec{x}}_{k}^{(n)}\}_{n=1}^{\tilde{N}_{k}}$ 
with associated RoI position 
$\{\vec{p}^{(i)}_{k}\}_{i=1}^{\tilde{N}_{k}}$ into the target video-sequence}
\newline {Target video sequence ${\cal I}_t $}
\newline {$\alpha_p$: overlap threshold}
\ENSURE {Set $\{\tilde{\pi}^{(i)}_{k} \}_{i=1}^{\tilde{N}_{k}}$ of weights associated to samples}
\STATE  -----------------------------------------------------------
\FOR{ $i=1$ to $\tilde{N}_{k}$}
\STATE {$\tilde{\pi}^{(i)}_{k}\leftarrow 0$}
\STATE {\tt\small /* Visual contextual cue computation */} 
\STATE {$\lambda_o =\dfrac{2(RoI\_AR\ \times\ FG\_AR)}{RoI\_AR + FG\_AR}$}
\STATE {\tt\small /* Weight assignment */} 
\IF {($\lambda_o \geqslant \alpha_p)$}
\STATE $\tilde{\pi}^{(i)}_{k} \leftarrow \lambda_o$
\ENDIF
\ENDFOR
\end{algorithmic}
\end{algorithm}

Considering the likelihood function, the favoring of sample associated to the right label becomes efficient and easier.

\subsection{\textbf{Fine-tuning step}}

In the proposed framework, the aim of the fine-tuning step is to specialize the RPN and the Fast R-CNN deep networks to a specific scene. Accordingly, we use the target detection boxes included in the specialized dataset ${\cal D}_{k}$ and the RPN fine-tuning process mentioned in \cite{DBLP:journals/corr/RenHG015}.

\begin{figure*}
\centering
\label{img4}
\centering
\includegraphics[width=1.5\columnwidth]{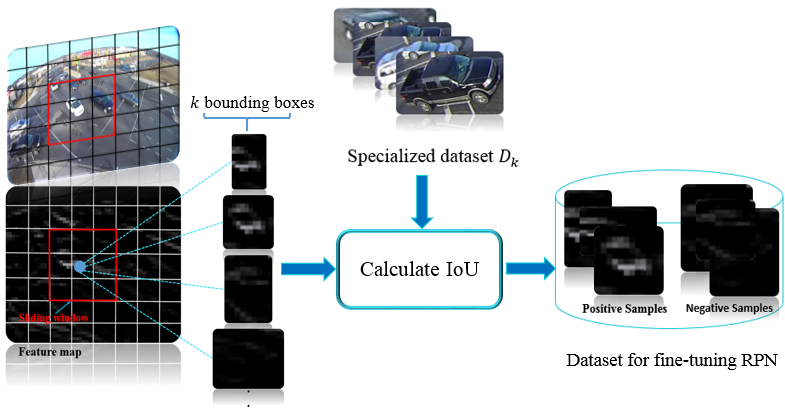}
\caption{\label{fig:fine}Description of training strategy for the RPN fully-convolutional network }
\end{figure*}

To do this, we use a sliding window approach to generate $k$ bounding boxes for each position on the feature map produced by the last convolutional layer, where each bounding box is centered on the sliding window and is associated with an aspect ratio and a scale (see Figure \ref{fig:fine}). The intersection-over-Union (IoU) overlap between each box of the specialized dataset ${\cal D}_{k}$ and the bounding boxes is then computed. 
A bounding box is designated as a positive training example if it has an IoU overlap greater than a predefined threshold with any ${\cal D}_{k}$ box, 
or if it is the bounding box that has the highest IoU with a ${\cal D}_{k}$ box. 
A proposal is designated as a negative example to a non-positive bounding box if its maximum IoU ratio with all boxes of the specialized dataset ${\cal D}_{k}$ is less than another predefined threshold. The bounding boxes that are neither positive nor negative do not contribute to the training.

Note that, the RPN fine-tuning process mentioned above does not consider that there might exist multiple copies (maximum twice) of the target detection box in the specialized dataset ${\cal D}_{k}$ because the main objective of using the IR algorithm proposed in the sampling strategy is not to increase the size of the database with samples which have high weights but to decrease the risk of distorting the specialized dataset ${\cal D}_{k}$ with wrong labelled examples because it is possible that the weighted target dataset contains wrong samples classified as traffic objects because their $\lambda_o$ $ >=$ $\alpha_p$. 

After training the RPN, these proposals are used to train the Fast R-CNN. Figure \ref{fig:fine} illustrates the training strategy of the RPN fully-convolutional network.

\begin{algorithm}
		\caption{ SMC Faster R-CNN}
		\label{alg1}
		\begin{algorithmic}
			\REQUIRE { Generic  network $({\cal {R}}_0,{\cal {F}}_0)$}
			\newline { Number of iterations: $K$}
			\newline { Number of target samples $\tilde{N}_{k}$}
			\newline { Unweighted target dataset: $\tilde{\cal W}_k$}
			\newline { Target video sequence ${\cal I}_t$ }
			\ENSURE  { Specialized network $({\cal {R}}_K,{\cal {F}}_K)$}
			\newline { Specialized dataset ${\cal D}_K$}
			\STATE  -----------------------------------------------------------------------
			\FOR{\texttt{k=1,...,K}}
			\STATE {{\tt\small /* Prediction step */}} 
			\STATE{$\{\tilde{\vec{x}}_{k}^{(n)}\}_{n=1}^{\tilde{N}_{k}} = \Theta(\{\mat{I}^{(i)}\}_{i=1}^{I_i};{\cal {R}}_{k-1},{\cal {F}}_{k-1})$}
			\STATE {{\tt\small /* Update step */}} 
			\STATE {$\tilde{\cal W}_k$ = $\{\tilde{\vec{x}}^{(n)}_{k},\tilde{\pi}^{(n)}_{k} \}_{n=1}^{\tilde{N}_k}$}
			\STATE {\tt\small /* Sampling step */} 
			\FOR{ $n=1$ to $\tilde{N}_{k}$}
			\STATE{Draw a sample: 
			$\{\tilde{\vec{x}}_{k}^{(n)}\}_{n=1}^{\tilde{N}_{k}}$
			, according to the weight}
			${\tilde{\pi}^{(n)}_{k}}$
			\ENDFOR
			\STATE {\tt\small /* Fine-tuning step */}
			\STATE {$\{{\cal {R}}_{k},{\cal {F}}_{k}\} =
			 f({\cal I}_t,{\cal D}_k;{\cal {R}}_{k-1},{\cal {F}}_{k-1})$
			}
						\ENDFOR
			\STATE {$\{{\cal {R}}_{K},{\cal {F}}_{K}\}
			=({\cal {R}}_{k},{\cal {F}}_{k})$ }
		\end{algorithmic}
	\end{algorithm}

Therefore, a new specialized RPN network and the Fast R-CNN one are generated being fine-tuning with the specialized dataset. These networks will become the input of the prediction step in the next iteration and will generate new object proposals (bounding boxes) in the target scene. 
\begin{equation}
	\{{\cal {R}}_{k},{\cal {F}}_{k}\} = f({\cal I}_t,{\cal D}_k;{\cal {R}}_{k-1},{\cal {F}}_{k-1})
\end{equation}

The suggested SMC Faster R-CNN framework is summarized in Algorithm \ref{alg1}.

\section{Experimental results}
This section presents the experiments that have been achieved in order to compare the SMC Faster R-CNN with the relevant frameworks on several public and private datasets for single and multi-traffic object detection.
\subsection{\textbf {Implementation details}}
We describe the implementation details of the SMC {Faster R-CNN} algorithm. 
We use the pre-trained VGG16 model \cite{simonyan2014very} to initialize the Faster R-CNN network, which is used in most recent state-of-the-art approaches \cite{girshick2015fast}\cite{Girshick_2014_CVPR}.

Both RPN and Fast R-CNN are fine-tuned end-to-end by back-propagation and stochastic gradient descent \cite{lecun1989backpropagation} with a weight decay of 0.0005 and a momentum of 0.9. We use the alternating training algorithm \cite{DBLP:journals/corr/RenHG015} for Faster R-CNN training from one iteration to another. The Faster R-CNN is fine-tuned on a NVIDIA GeForce GTX TITAN X GPU with a 12GB memory. 

Following multiple experiments, we 
chose 9 as the number of bounding boxes (3 aspect ratios [2:1, 1:1, 1:2] and 3 scales [$128^2$, $256^2$, $512^2$]) generated on each position of the sliding windows. We also chose 0.7 as the threshold of the IoU to select the positive samples and 0.3 for the negatives to build the training dataset.

\begin{figure}
\label{imgit}
\includegraphics[width=\columnwidth]{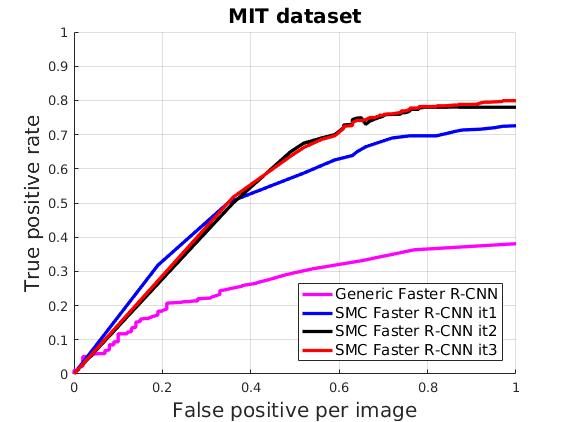}
\caption{\label{fig:fine13}ROC curves for convergence of specialization process}
\end{figure}

The parameter $K$  (number of iterations of the SMC process) is fixed to $K=2$. Figure \ref{fig:fine13} shows that the specialization converges after two iterations for both car and pedestrian detection applied on the MIT Traffic dataset (introduced in the next section).



\subsection{\textbf {Datasets}}
\label{dataset}
\begin{figure*}
\label{img1}
\centering
\includegraphics[width=1.2\columnwidth]{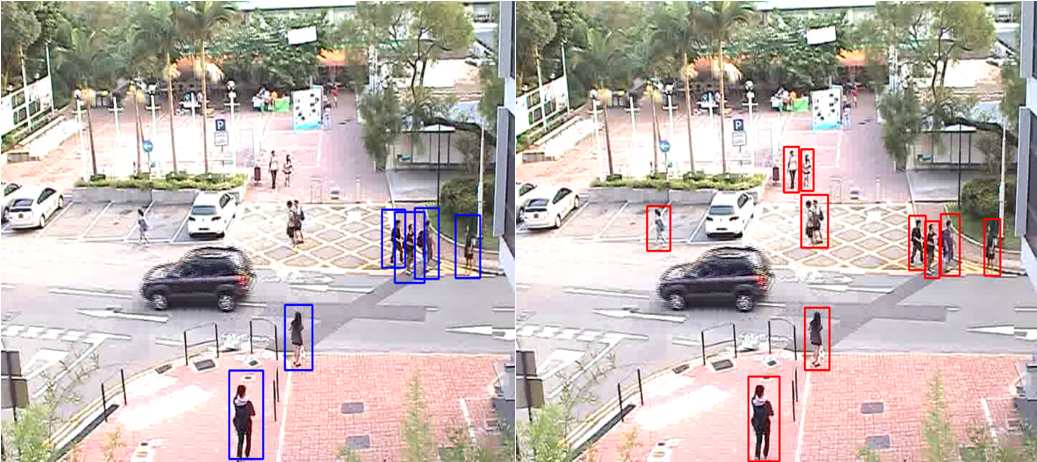}
\\
(a)\\
 
\includegraphics[width=1.2\columnwidth]{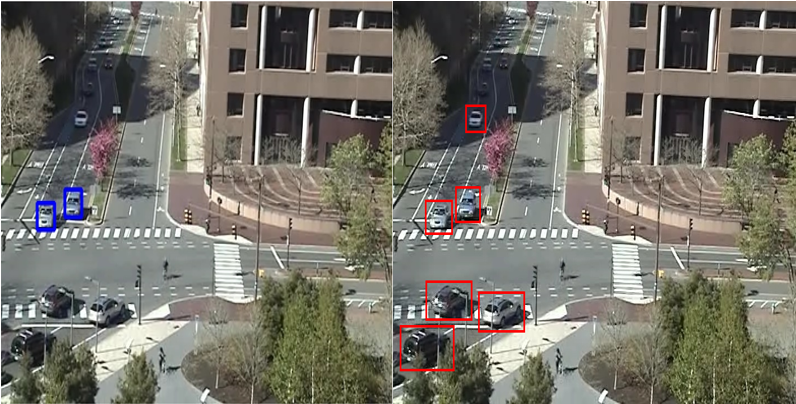}\\
(b)\\
\caption{\label{fig:gt}
Some annotations provided by Wang \cite{wang2014scene} ground truth on CUHK dataset ((a) left image), Li \cite{li2015domain} ground truth on MIT Traffic dataset ((b) left image) and our annotations ((a), (b) right images). There are several missing objects in the baseline annotations: This is why we propose an updated version.} 
\end{figure*}

The PASCAL VOC 2007 dataset \cite{everingham2010pascal} was utilized to learn the generic Faster R-CNN. This dataset consists of about 5,011 trainval images and 4,952 test ones over 20 object categories. In our experiments, we use only 713 annotated cars, 2,008 pedestrian, 186 buses and 245 for motorbikes, to fine-tune the generic Faster R-CNN. The evaluation is achieved on three target datasets (two public ones and a private one):
\begin{itemize}
\item \textbf{CUHK Square dataset \cite{wang2012transferring}:}
This is a video sequence of road traffic
which lasts 60 minutes. 352 images are utilized for specialization, uniformly extracted from the first half of the video.
100 images are used for the test, extracted from the latest 30 minutes. Annotations were provided by Wang \cite{wang2012transferring} for pedestrian detection (called \textbf{CUHK\_WP} after). However, we notice that some annotation errors are made in the public ground truth and we suggest a new annotation (called \textbf{CUHK\_MP} after) (see \figurename~\ref{fig:gt}.a). 
\item \textbf{MIT Traffic dataset \cite{wang2009unsupervised}:} This is a 90-minute video. We use 420 images from the first 45 minutes for specialization. 100 images are uniformly sampled from the last 45 minutes for the test. Annotations are available for pedestrians \cite{wang2009unsupervised} (called \textbf{MIT\_WP}) and cars \cite{li2015domain} (called \textbf{MIT\_LV)}. Since some annotation errors are present, we propose new annotations (called \textbf{MIT\_MV}) (see \figurename~\ref{fig:gt}.b).
\item \textbf{{Logiroad Traffic dataset:}} This is a private video sequence of road traffic which lasts 20 minutes. We utilize 600 images for specialization, extracted uniformly from the first 15 minutes of the video. 100 images are used for the test, extracted from the latest 5 minutes. Annotations are available for vehicles (called \textbf{Logiroad\_MV}).
\end{itemize}

\subsection{\textbf {Descriptions of experiments}}

Evaluation is performed in terms of recall False Positives Per Image (FPPI) curves. The PASCAL 50 percent overlap criterion \cite{everingham2010pascal} was utilized to give a score for the detection bounding boxes. 
The SMC Faster R-CNN framework is compared with several state-of-the-art frameworks:
\begin{itemize}
\item Generic {Faster R-CNN}: It is a detector fine-tuned on the generic dataset. This is the baseline for our comparison.
\item Maamatou (2016) \cite{maamatou2016}: An SMC framework was applied to specialize a generic HOG-SVM classifier to a particular video sequence for traffic object detection.

\item Xudong Li (2015) \cite{li2015domain}: A deep learning domain adaptation framework was proposed for vehicle detection with manually annotated data from the target scene. Unlike other methods, the latter was not totally automatic and requires some manual annotations.

\item Mao (2015) \cite{mao2015}: A framework was suggested to automatically train scene-specific pedestrian detectors based on tracklets.

\item Htike (2014) \cite{htike2014efficient}: A non-iterative domain adaptation framework was used to adapt a pedestrian detector to video scenes. 

\item Zeng (2014) \cite{zeng2014deep}: A deep learning domain adaptation framework was proposed to automatically select training samples from target scenes without manual labelling for pedestrian detection.

\item Wang (2014) \cite{wang2014scene}: A specific-scene detector was trained on only relevant samples collected from both source and target datasets.

\item Nair (2004) \cite{nair2004unsupervised}: An iterative self-training framework for detector adaptation was opted for using a background subtraction algorithm.

\end{itemize}

\subsection {\textbf {Results and analysis for single-traffic object}}
Given each dataset and its annotation, we present the ROC curves (Figure \ref{fig:fine12}) of the generic Faster R-CNN, the SMC Faster R-CNN and the available state-of-the-art frameworks. The ROC curves present the comparison between the true detection rate and the false positive detection rate per image. Furthermore, we give two comparative synthetic tables: one for pedestrian detection (cf. \tablename~\ref{timep}) and the other for vehicle detection (cf. \tablename~\ref{timep2}). In addition, on the last line of both tables, the improvement between the generic Faster R-CNN and the SMC {Faster R-CNN} is given. 
\begin{itemize}

\item \textbf{Comparison with generic detector:}
Figure \ref{fig:fine12} shows that the specialized Faster R-CNN detector significantly outperforms the generic one on all public and private datasets with several annotations. The median improvement is \textbf{$51\%$}.


\item \textbf{Comparison with state-of-the-art:}
According to the ROC curves at the top of Figure \ref{fig:fine12}, for the CUHK pedestrian detection, the SMC Faster R-CNN outperforms all other state-of-the-art frameworks. Besides, the detection rate achieved with our proposed annotations on \textbf{CUHK\_MP} is nearly \textbf{$90\%$} for 0.5 FPPI.
However, despite of the wrong annotations given by Wang (left curve in the top of Figure \ref{fig:fine12}), the SMC Faster R-CNN also exceeds the six other specialized detectors of Nair (2004), Wang (2014), Zeng (2014), Htike (2014), Mao (2015) and Maamatou (2016) respectively by 24\%, 45\%, 53\%, 49\%, 58\% and 62\%.

For the MIT pedestrian detection (\textbf{MIT\_WP} in Table 2), the specialized deep detector proposed by Zeng (2014) exceeds the SMC Faster R-CNN detector for an 0.5 FPPI, which is less than 0.9.

Despite the wrong annotations given by Li  {\it {\it {\it et al.}}} \cite{li2015domain}, Figure \ref{fig:fine12} (right curve in the middle) shows that for the MIT car detection (\textbf{MIT\_LV}), the proposed SMC Faster R-CNN clearly outperforms the specialized CNN detector proposed by Li (2015) which trained with manual data labeling from the target scene.
According to Table 3, for the MIT and Logiroad car detection with the proposed annotations, the SMC Faster R-CNN is ranked first and exceeds the specialized detector suggested by Maamatou (2016).

One can notice that the generic Faster R-CNN, fine-tuned on the PASCAL VOC 2007 dataset, has a poor detection rate resulting in a limitation of the size of the specialized dataset.
\begin{table}[!ht]
\caption{\label{timep} Comparison of detection rate for pedestrian with state of the art (at 0.5 FPPI)}
\small
\begin{tabular}{l||l|l|l}
\small
\backslashbox{Approach}{Dataset} & CUHK\_WP& CUHK\_MP & MIT\_WP\\
\hline 
Nair \cite{nair2004unsupervised}  & 0.24 & -- & 0.35\\
Wang \cite{wang2014scene} & 0.45 & -- & 0.42\\
Zeng \cite{zeng2014deep} & 0.53 & -- & 0.58\\
Htike \cite{htike2014efficient} & 0.49 & -- & -- \\ 
MAO  \cite{mao2015} & 0.58 & -- & -- \\ 
Maamatou \cite{maamatou2016} & 0.62 & 0.58 & 0.40 \\
Generic {Faster R-CNN} \cite{DBLP:journals/corr/RenHG015} & 0.60 & 0.69 &0.07\\
SMC {Faster R-CNN} & \textbf{0.65} & \textbf{0.88} & \textbf{0.47} \\
\hline
Improvement / generic (\%)& 8\% & 28\% & 571\% \\
\end{tabular}
\end{table}
\small
\begin{table}[!ht]
\small
\caption{\label{timep2} Comparison of detection rate for car with state of the art (at 1 FPPI)}
\begin{tabular}{l||l|l|l}

\backslashbox{Approach}{Dataset} & MIT\_LV & MIT\_MV & Logiroad\_MV  \\
\hline
Li \cite{li2015domain} & 0.77 & --& --\\ 
Maamatou \cite{maamatou2016} & -- & 0.29 & 0.47  \\
Generic {Faster R-CNN}  \cite{DBLP:journals/corr/RenHG015} & 0.68 & 0.38 & 0.40 \\
SMC {Faster R-CNN} & \textbf{0.77} & \textbf{0.80} & \textbf{0.70} \\
\hline
Improvement / generic (\%) & 13\% & 110\%  & 75\% \\
\end{tabular}
\end{table}

\begin{figure*}
\label{img111}
\includegraphics[width=\columnwidth]{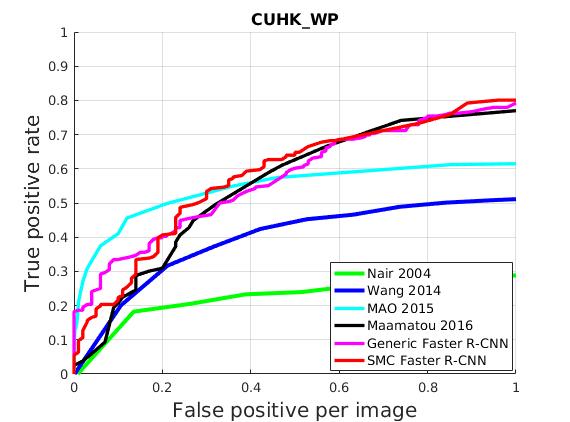}
\includegraphics[width=\columnwidth]{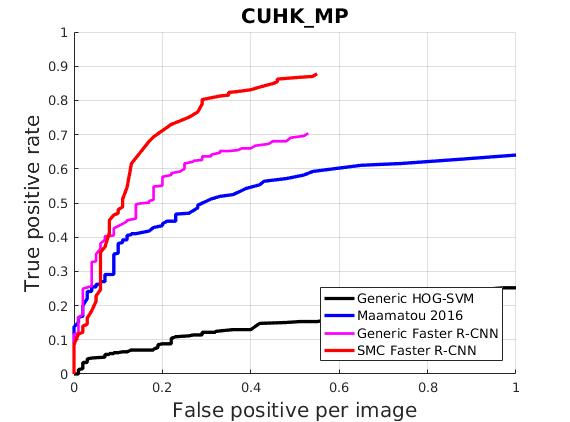}
\includegraphics[width=\columnwidth]{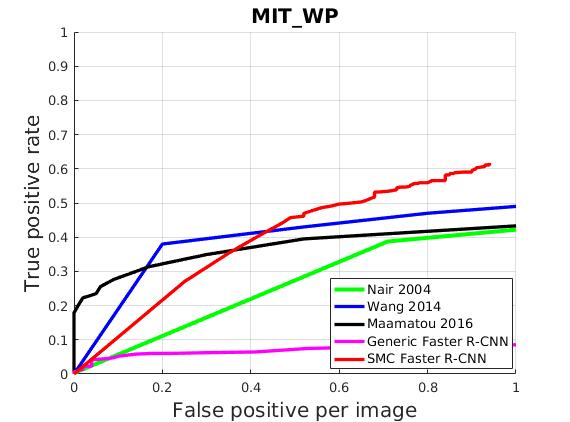} 	
\includegraphics[width=\columnwidth]{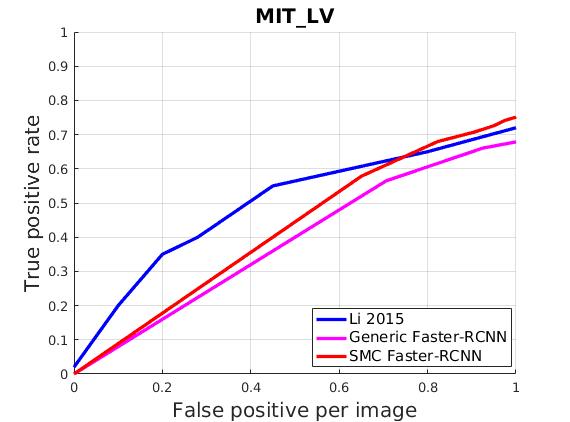} 
\includegraphics[width=\columnwidth]{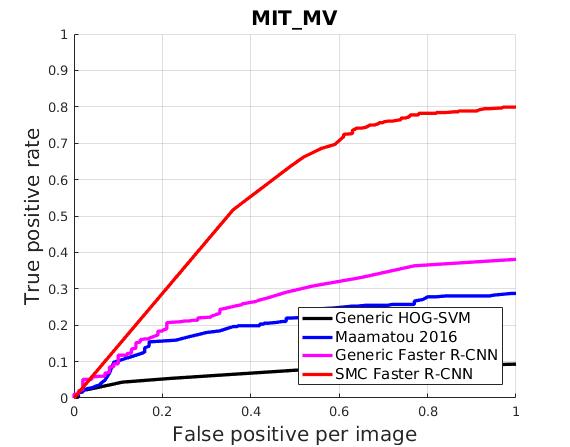}
     \includegraphics[width=\columnwidth]{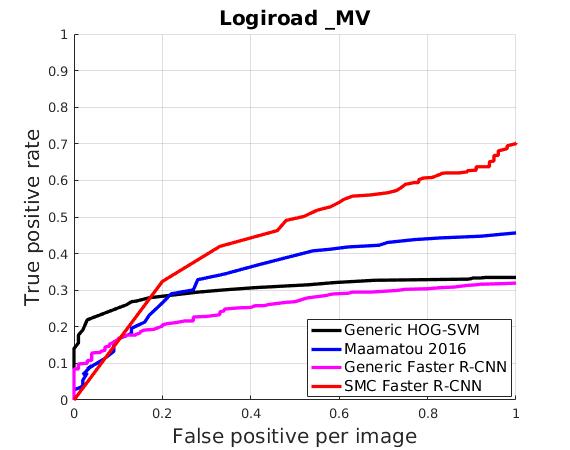} 
     \caption{\label{fig:fine12}ROC curves for several datasets and annotations}
\end{figure*}
\end{itemize}
 
\begin{figure*}
\label{imgit}
\centering
\includegraphics[width=1.6\columnwidth]{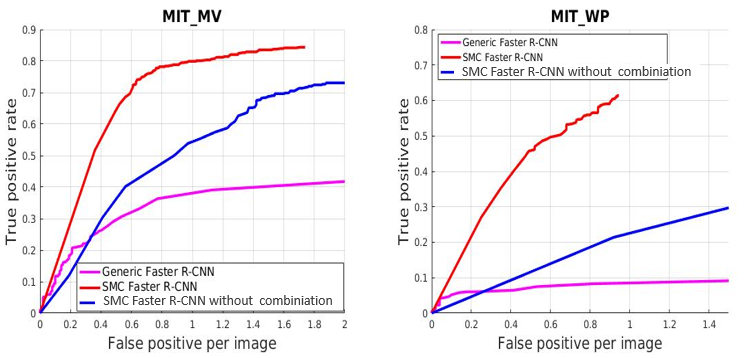}

\caption{\label{fig:fine14}Effect of the likelihood function in our specialization framework on MIT Traffic dataset for both pedestrian and car detections.
}
\end{figure*}

\begin{itemize}
\item \textbf{Effect of likelihood function:}
To show the effectiveness of our likelihood function, the ROC curves in Figure \ref{fig:fine14} show the comparison between using the likelihood function based only on confidence score predicted by the Faster R-CNN and our proposed one on two datasets.

The red curves in Figure \ref{fig:fine14} present our proposed likelihood function based on the combination between the confidence score and the spatio-temporal cue, and the blue ones indicate the use of the confidence score only, which is given by the output layer of the Faster R-CNN. 
The results demonstrate that the proposed likelihood function based on using the verification strategy improves the detector performance and accelerates the convergence of the specialization process. Furthermore, we cannot say that this choice is the best because it is possible to ameliorate the suggested framework by proposing other strategies for the SMC steps. For example, we can improve the likelihood function with more complex visual cues like tracking, optical flow or contextual information to enhance the weighting of positive samples.
\end{itemize}

\subsection {\textbf {Results and analysis for multi-traffic object}}

We evaluate the proposed approach for multi-traffic objects on two datasets, the MIT Traffic dataset and the Logiroad one using two evaluation criteria: namely the ROC curves and the confusion matrix (classical metrics for object detection).

For the MIT Traffic dataset, we select 2 classes \{'pedestrian', 'car'\} and 4 classes for the Logiroad Traffic dataset \{'pedestrian', 'car', 'bus', 'motorbike'\}.

The results are reported in \tablename~\ref{timep5}. The SMC Faster R-CNN presents a median improvement of $89\%$ related to the generic detector.
Moreover, Tables \ref{timep30} and \ref{timep40} provide the associated similarity matrix. We show that some confusion may occur between motorbikes and cars or between buses and cars. Furthermore, these results illustrate that our framework has a robust performance for multi-traffic object detection.
This indicates that it is useful to run our specialization algorithm whenever we have a new sequence and we want to automatically generate a much better deep detector than the generic one.

\begin{table*}[!ht]
\caption{\label{timep5} Detection rate for multi-traffic object detection with SMC {Faster R-CNN} (at 1 FPPI)}
\centering
\begin{tabular}{l||l|l|l|l|l}
\centering
\backslashbox{Approach}{Dataset} & {Logiroad\_Car} & {Logiroad\_Person}&{Logiroad\_Moto} & {MIT\_Car} & {MIT\_Person}\\
\hline
 Generic Faster R-CNN & {0.28} & {0.24}& 0,065 & {0.32} & {0.05} \\
SMC {Faster R-CNN} & \textbf{0.60} & \textbf{0.36} & \textbf{0.18} &\textbf{0.73} & \textbf{0.30} \\
 \hline
Improvement/ generic(\%) & 114\% & 50\% & 176\% & 128\% & 500\% \\
\end{tabular}
\end{table*}

\begin{table*}[!ht]
     \caption{\label{timep30}Illustration of similarity matrix between traffic object categories on Logiroad Traffic dataset (diagonal row shows the accuracy to recognize traffic objects of its own class)}
\centering
\begin{tabular}{l||l|l|l|l}
\centering
\backslashbox{Actual class}{Predicted class} & Pedestrian & Car & Motorbike & Bus \\
\hline
Pedestrian & 140/\textbf{97\%} & 12/1.5\% & 5/14\% & 0 \\ 
Car & 0 & 750/\textbf{96\%} & 1/3\% & 1/2.5\\
Motorbike & 5/3\% & 12/1.5\% & 30/\textbf{83\%} & 1/2.5\% \\
Bus & 0 & 7/1\% & 0 & 38/\textbf{95\%}\\
\hline
Total & 145 & 781 & 36 & 40\\
\end{tabular}
\end{table*}



\begin{table}[!ht]
    \caption{\label{timep40} Illustration of similarity matrix between traffic object categories on MIT Traffic dataset}
\centering
\begin{tabular}{l||l|l}
\centering
\backslashbox{Actual class }{Predicted class } & Pedestrian & Car \\
\hline
Pedestrian & 342/\textbf{99.7}\% & 7/1.6\% \\ 
Car & 1/0.3\% & 420/\textbf{98.4}\% \\
\hline
Total & 343 & 427 \\
\end{tabular}
\end{table}

\section{Conclusion and future work}

We have put forward an efficient framework based on the formalism of the SMC filter to specialize the Faster R-CNN deep detector for multi-traffic object detection. This framework approximates the unknown target distribution by selecting relevant samples from target datasets. These samples are utilized to fine-tune a specialized deep detector in order to decrease the detection rate in the target scene. Given a generic detector and a target video sequence, this framework automatically provides a robust specialized detector.
Moreover, the proposed framework allows reducing the risk of drifting by using efficient strategies during iterations and it can be used to specialize any deep detector. 
The extensive experiments have demonstrated that the suggested framework has produced a specialized detector that performs much better than the generic one for both single and multi-traffic object  detections in different scenes. Furthermore, the results show that the framework outperforms the state-of-the-art specialization ones on several challenging datasets. Our future work will deal with an extension of the algorithm to improve the likelihood function by using a new strategy of verification based on more complex visual cues like tracking, optical flow, tracklets or contextual information and injecting some spatio-temporal information into the Faster R-CNN network.




\section*{Acknowledgment}

This work is within the scope of a co-guardianship between the university of Sousse (Tunisia) and Clermont Auvergne University (France). It is sponsored by the Tunisian Ministry of Higher Education \& Scientific Research and the French government research program ''Investissements d'avenir'' through the IMobS3 Laboratory of Excellence (ANR-10-LABX-16-01), by the European Union through the program Regional competitiveness and employment 2007-2013 (ERDF - Auvergne region), and by the Auvergne region.

\section*{References}


\bibliography{mybibfile}

\end{document}